# A data balancing approach towards design of an expert system for Heart Disease Prediction

Rahul Karmakar[1] [https://orcid.org/0000-0002-6607-2707], Udita Ghosh[1], Arpita Pal[1], Sattwiki Dey[1], Debraj Malik[1] and Priyabrata Sain[1]

[1] The University of Burdwan, Bardhaman West Bengal, India

```
rkarmakar@cs.buruniv.ac.in,uditaghosh2001@gmail.com,
arpitapal1122@gmail.com, sattwikidey55@gmail.com,
debrajmalik08@gmail.com, papai25011992@gmail.com
```

**Abstract.** Heart disease is a serious global health issue that claims millions of lives every year. Early detection and precise prediction are critical to the prevention and successful treatment of heart-related issues. A lot of research utilises machine learning (ML) models to forecast cardiac disease and obtain early detection. In order to do predictive analysis on "Heart_disease_health_indicators " dataset .We employed five machine learning methods in this paper: Decision Tree (DT), Random Forest (RF), Linear Discriminant Analysis, Extra Tree Classifier, and AdaBoost. The model is further examined using various feature selection (FS) techniques. To enhance the baseline model, we have separately applied four FS techniques: Sequential Forward FS, Sequential Backward FS, Correlation Matrix, and Chi2. Lastly, K_means SMOTE oversampling is applied to the models to enable additional analysis.

The findings show that when it came to predicting heart disease, ensemble approaches—in particular, random forests—performed better than individual classifiers. The presence of smoking, blood pressure, cholesterol, and physical inactivity were among the major predictors that were found. The accuracy of the Random Forest and Decision Tree model was 99.83%. This paper demonstrates how machine learning models can improve the accuracy of heart disease prediction, especially when using ensemble methodologies. The models provide a more accurate risk assessment than traditional methods since they incorporate a large number of factors and complex algorithms.

# Introduction

As the most prevalent cause of death all over the world, cardiovascular disease (CVD) has gained significant attention as a public health issue. The governments of these countries, as well as patients and their families, have thus had to incur significant socioeconomic costs. Using risk stratification, prediction models can determine which people are more likely to develop cardiovascular disease. Once that risk has been reduced, population-specific interventions such as the use of statins and dietary changes can help to improve primary CVD prevention [1].

Cardiovascular and blood vessel diseases, or CVDs, classify illnesses like myocarditis, heart disease, and vascular disease. 80% of CVD patients die from heart disease and stroke. Seventy-five percent of all deceased people are under the age of seventy. Age, gender, drinking alcohol, high blood pressure, dietary fats, inactivity, and smoking are the primary factors in cardiovascular disease. Over 17.8 million people die of heart disease directly every year, according to a WHO report. One of the leading causes of death is heart disease, followed by stroke [2]. Predictive models have been recommended by multiple guidelines for the assessment and management of CVD as a way to identify high-risk patients and support clinical decision-making. [3].

Electrocardiograms and CT scans, for example, are just two of the diagnostic instruments that are too expensive and complex for the average person to use to diagnose coronary heart disease. Just the aforementioned cause has claimed the lives of 17 million people [4]. Workers who suffered from cardiovascular diseases were responsible for 25–30% of the companies' annual medical expenses. Consequently, early identification is essential to reducing the financial and medical burden that cardiovascular disease places on individuals and institutions[5].

According to World Health Organization predictions, heart disease and stroke will be the primary causes of the 23.6 million deaths caused by CVD that occur worldwide by 2030 [6]. Due to its difficult diagnostic process, heart failure has been the focus of much research [7]. For this reason, computer-assisted decision support systems, like the one described in [8], are very helpful in this area. Data mining techniques were used to shorten the time required to make an accurate prediction of the disease.

This work focusses on a comprehensive analysis of heart disease prognosis. Over the years, ML algorithms have been employed extensively in this field. After conducting a thorough performance analysis of our suggested models with varying parameters, we suggest an improved machine learning model for the prediction of heart disease.

### 1.1    Contribution of the paper

The main focus of this research is to examine the predictive capabilities of machine learning and ensemble techniques in heart disease diagnosis. The underlying model is constructed using five machine learning methods, and the best result is evaluated using ten-fold Cross-Validation. Then, in order to enhance the underlying model, we employed feature selection approaches. Ultimately, the data balancing method is utilized to construct a final model. We have methodically examined every performance in every scenario.

### 1.2    Structure of the paper

The organization of the paper as follows: In Section 1, we have described the introduction of machine learning as well as heart disease. Section 2 we have done the literature review. Section 3 represents the case study and data pre-processing. Section 4 discussed the result and performance analysis. The comparative study demonstrated in section 5 followed by Conclusion in section 6.

## 2 Literature Review

The goal of the literature review is to give researchers and healthcare professionals a thorough understanding of the state-of-the-art in heart disease prediction using machine learning techniques.

In a CHD prediction study, the authors proposed many classification methods including AUC, RF, LR, SVM, XG Boost, eXtreme Gradient Boost, and Light GBM. The study recruited 7672 individuals aged 30 to 84 without cardiovascular disease, monitored them for an average of 15 years, and identified key predictors of CHD, including systolic blood pressure, non-HDL-c, glucose levels, age, metabolic syndrome, HDL-c, estimated glomerular filtration rate, hypertension, elbow joint thickness, and diastolic blood pressure [9].

Daniyal Asif conducted a study in 2023 using Extra Tree Classifier, Random Forest, CatBoost, and XGBoost classifiers to predict heart illness. Using an ensemble learning model, the researchers added an extra tree classifier, trained and tested using an 80:20 split, and optimized hyperparameters using grid search cross-validation. The extra tree classifier generated excellent results, with 97.23% accuracy, 98.72% recall, 95.68% precision, and a 97.18% F1 score [10].

In 2022, in a study conducted by Muhammad Salman Pathan, the CVD dataset was collected and various ML models were used, like SVC, LR, AdaBoost, LGBM, Extra Tree Classifier, XGB Classifier, RF, Gaussian NB, etc. Considering the classification results, the best accuracy reported was 75% using the SVC classifier. Furthermore, it achieves a ROC value of 74% and an F1 score of 73% [11].

In Shu Jiang's 2020 study, with an astounding accuracy score of 88.5%, the Random Forest classifier surpassed previous machine learning methods. With accuracy scores of 83.6, 85.2, and 85.2, respectively, XG Boost, Logistic Regression, and Neural Network are the other techniques implemented [12].

Similarly, in 2021, Hisham Khdair and Naga M Dasari used a variety of classification algorithms to predict heart illness, including SVM, MLP Neural Network, KNN, and Logistic Regression. Out of all the classification models, the support vector machine (SVM) performed the best, having an accuracy of 73.8% [13].In 2019, a study conducted by S. Mohan and others, the UCI Cleveland dataset was collected and various ML models were used like Naïve Bayes, LR, DT, SVM, Deep learning, VOTE, RF, HRFLM, etc. Considering the classification results the best accuracy reported was 88.4% using the HRFLM classifier [14].In a study published in 2021, Harshit Jindal and colleagues used Random Forest, Logistic Regression, and KNN classifiers to predict cardiovascular illness. With an accuracy of 88.52%, the KNN classifier did well [15]. Pooja Rani and colleagues utilized classifiers such as Random Forest, Logistic Regression, SVM, Naïve Bayes, and AdaBoost in their research. The Random Forest classifier scored well, with 86.60% accuracy, 84.14% sensitivity, 88.46% precision, an F-measure score of 97.18%, and 89.02% specificity [16].

Ten primary algorithms SVM, RF, LR, DTs, NB, Generalized Linear Model, Deep Learning, Gradient Boosted Trees, VOTE, and HRFLM were used in a study on the heart disease dataset to compute, compare, and assess various outcomes depending on the F-measure, accuracy, sensitivity, specificity, and precision [17].

## 2.1   Research Gaps

The majority of research does not address the issue of class imbalance, the feature selection approaches are not used systematically, and the heart disease dataset is improperly pre-processed. It is also worth noting that some of these systems do not use cross-validation, a critical stage in performance validation. We addressed every issue raised in your proposal and enhanced the accuracy of the prognosis for heart disease.

## 3   A case study on Heart Disease Prediction

In this section we have done a case study on heart disease prediction using Machine Learning. The following from (3.1 to 3.6) subsection demonstrated the overall operation. The overall flow diagram is represented using Figure 1.

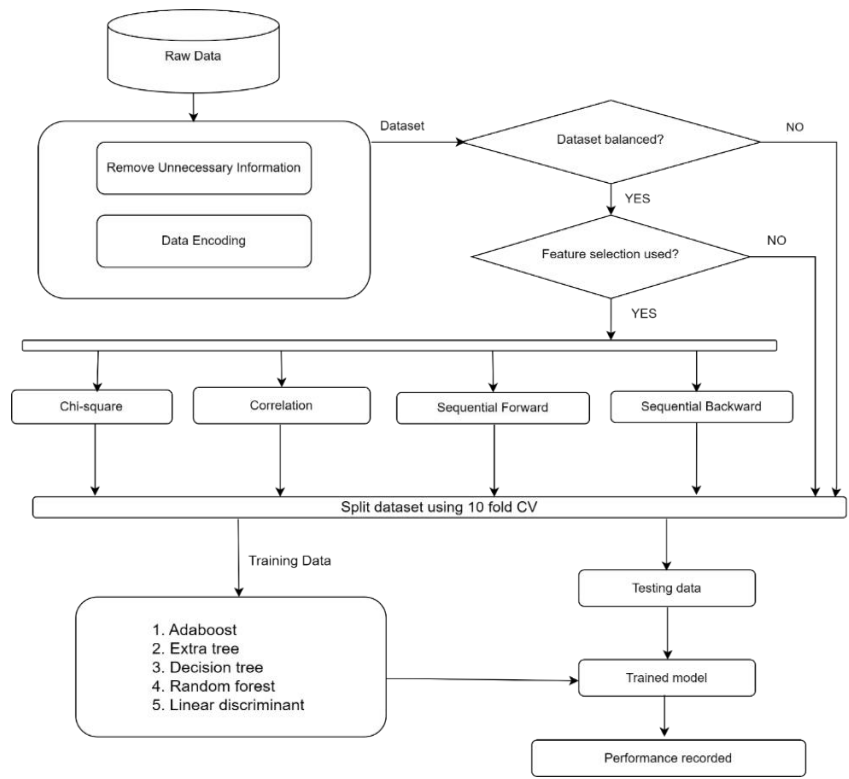

**Figure 1**: Flow diagram of case study

**3.1    Dataset Description**

The dataset "Heart_disease_health_indicators" is used in this investigation. It acts as a foundational cornerstone for developing and evaluating predictive models aimed at discerning patterns, elucidating disease mechanisms, and predicting clinical outcomes in individuals afflicted with heart disease. The dataset used in this work comes from a collection of biomedical datasets that have been carefully selected and made accessible to the public for scientific use.
We have taken the dataset from Kaggle for heart disease analysis. This dataset is made up of 100000 instances and 22 features. The dataset contains 9340 positive cases and 90660 negative cases instances. These two classes are assigned "1" and "0" labels respectively. No missing value or null value is present in the dataset.

**3.2    Data Pre-Processing**

A critical phase in the machine-learning process is data preprocessing, which entails arranging, modifying, purifying, and disinfecting raw data to prepare it for use in machine-learning model training. Errors like missing values, outliers, and inconsistencies are removed to enhance the data's quality and utility. Additionally, the data is ready for machine learning algorithms to use efficiently. Data cleaning deals with incorrect or missing data handling. Missing data may be addressed by methods like interpolation, imputation (replacing missing values with estimated values), and record deletion of incomplete data.

**3.3    Data balancing using K-Means SMOTE oversampling**

Preparing a model is facilitated by balancing a dataset, as it helps keep the model from becoming biased in favor of one class. It helps to prevent algorithms from assigning all cases to the most common outcome and can improve accuracy rates. Balanced data sets are imperative in a data storage system to keep up reliable performance and prevent non-consistency in data access command execution. The balanced dataset has 181338 instances and 22 features. The balanced dataset contains 90678 positive cases and 90660 negative cases instances. These two classes are assigned "1" and "0" labels respectively.

**3.4    Cross Validation**

Cross-validation is a widely used technique to determine which machine-learning model is best. The procedure comprises fold division of the dataset, model training on one of the folds, and evaluation of the model on the remaining fold or folds. The final evaluation parameter is the average performance over all folds, and this procedure is iterated multiple times

## 3.5    Feature Selection

Feature selection, variable selection, or attribute selection is the process of choosing from a bigger collection a subset of relevant characteristics or variables to use in predictive models. It enhances the model's interpretability, decreases overfitting, and boosts accuracy. The basic goal of feature selection is to improve model performance by emphasizing the most valuable traits and removing the unneeded or unwanted ones. Feature selection is essential for determining the most important characteristics that improve a model's predictive power.

## 3.6    Model Evaluation

In the realm of AD research, machine learning algorithms in particular have attracted a lot of attention because of their capacity to learn from data and generate predictions devoid of explicit programming. These algorithms can leverage large-scale datasets to develop predictive models for AD risk assessment, disease progression monitoring, and treatment response prediction. By analyzing diverse features extracted from neuroimaging scans, genetic variants, and clinical records, machine-learning algorithms can identify subtle patterns and associations that may elude traditional analytical approaches.

## 4    Result and Performance Analysis

In this section, we have discussed the result and the performance analysis. Here we employed five machine learning stated algorithms to find the best result.

### 4.1    Results using a 10-fold CV without Feature Selection and without Data Balancing

In terms of performance, the Extra Tree classifier outperformed the other methods in the 10-fold CV in terms of precision, regarding accuracy and recall the Decision Tree technique, and the Adaboost classifier in terms of f1_score. The Decision Tree algorithm, which used the 10-fold CV method, had the highest accuracy of any strategy, at 99.73 percent.

### 4.2    Results using a 10-fold CV without Feature Selection and with Data Balancing

In terms of performance, the Random Forest classifier outperformed alternative approaches in the 10-fold CV based on f1_score and precision, while the Decision Tree technique outperformed alternative approaches based on accuracy and recall. The Decision Tree algorithm using the 10-fold CV method was the most accurate of all the approaches, with a 99.83 percent accuracy rate.

### 4.3    Results using a 10-fold CV with $Chi^2$ Feature Selection and without Data Balancing

.Performance wise, the Random Forest and Decision Tree techniques outperformed the others in terms of accuracy, the Linear Discriminant classifier outperformed the others in terms of recall, and the Extra Tree classifier outperformed the others concerning accuracy and f1-score in the 10-fold CV. By using the 10-fold CV method, the Decision Tree and Random Forest algorithm achieved the highest accuracy rate of all the strategies, at 91.71 percent.

### 4.4 Results using a 10-fold CV with Correlation Feature Selection and without Data Balancing

The Extra Tree classifier performs better than other methods in the 10-fold CV in terms of precision, the Random Forest and Decision Tree methods concerning recall and accuracy, and the AdaBoost classifier in terms of f1_score. The Decision Tree algorithm, which used the 10-fold CV method, had the highest accuracy of any strategy, at 99.47 percent.

### 4.5 Results using a 10-fold CV with SFFS Technique and without Data Balancing

The Random Forest strategy fared better in terms of precision, the Extra Tree classifier in terms of accuracy and recall in the 10-fold CV, and the Linear Discriminant classifier in terms of f1 score. With an accuracy of 90.88 percent, the Extra Tree algorithm, which employed the 10-fold CV method, had the highest of all the strategies.

### 4.6 Results using a 10-fold CV with SBFS Technique and without Data Balancing

The Random Forest technique fared better in terms of f1_score, the Extra Tree classifier did better in terms of accuracy in the 10-fold CV, and the Decision Tree and AdaBoost classifiers performed better in terms of recall and precision. With an accuracy of 90.87 percent, the Extra Tree algorithm, which employed the 10-fold CV method, had the highest accuracy of all the methods.

### 4.7 Results using a 10-fold CV with $Chi^2$ Feature Selection and with Data Balancing

The Decision Tree methodology outperformed the other approaches in terms of f1-score and precision in the 10-fold CV, but the Random Forest classifier did better concerning accuracy and recall. The most accurate approach among all the strategies was the Random Forest algorithm, which employed the 10-fold CV method, with a score of 82.51 percent.

### 4.8 Results using a 10-fold CV with Correlation Feature Selection and with Data Balancing

Performance-wise, the Random Forest classifier beat the other methods in the 10-fold CV for accuracy, f1_score, and precision, while the Decision Tree method beat the others for recall and accuracy. With an accuracy of 98.90 percent, the Decision Tree and Random Forest algorithm, which employed the 10-fold CV method, had the best accuracy of any strategy.

### 4.9 Results using a 10-fold CV with SFFS Technique and with Data Balancing

When it came to f1-score and precision in the 10-fold CV, compared to the other techniques, the Random Forest classifier performed better, while the Decision Tree methodology performed better in terms of recall and accuracy. Among all strategies, the Decision Tree algorithm, which employed the 10-fold CV method, had the best accuracy, at 90.18 percent.

**4.10   Results using a 10-fold CV with SBFS Technique and with Data Balancing**

The Random Forest classifier beat the other methods in the 10-fold CV regarding precision and f1-score, however, the Decision Tree algorithm outperformed them regarding recall and accuracy. With a rate of 90.69 percent, the Decision Tree algorithm, which employed the 10-fold CV method, was the most precise of the techniques.

Here we analyzed various metrics like accuracy, f1 score, precision, and recall.

Table 1 represent the performance analysis for accuracy metrics

**Table 1.** Performance analysis of Accuracy

| ML Classifiers | Without FS &Data Balancing | With Data Balancing | $Chi^2$ (Without Data Balancing) | Correlation(Without Data Balancing) | SFFS(Without Data Balancing) | SBFS (Without Data Balancing) | $Chi^2$ | Correlation | SFFS | SBFS |
|---|---|---|---|---|---|---|---|---|---|---|
| DT | 99.73 | **99.83** | 91.71 | 99.47 | 89.00 | 89.13 | 82.50 | 98.90 | 90.18 | 90.69 |
| RF | 99.73 | **99.83** | 91.71 | 99.46 | 90.81 | 90.81 | 82.51 | 98.90 | 89.67 | 87.59 |
| LDA | 90.27 | 78.90 | 89.93 | 90.17 | 90.68 | 90.70 | 73.57 | 76.88 | 78.76 | 78.84 |
| Extra tree | 90.66 | 74.08 | 90.60 | 90.60 | 90.88 | 90.87 | 70.15 | 73.37 | 82.86 | 84.60 |
| AdaBoost | 90.76 | 82.59 | 90.62 | 90.67 | 90.82 | 90.81 | 76.69 | 80.05 | 82.09 | 82.05 |

Figure 2 illustrates the accuracy of different classifiers with data balancing, without data balancing, along with feature selection methods, or without feature selection.

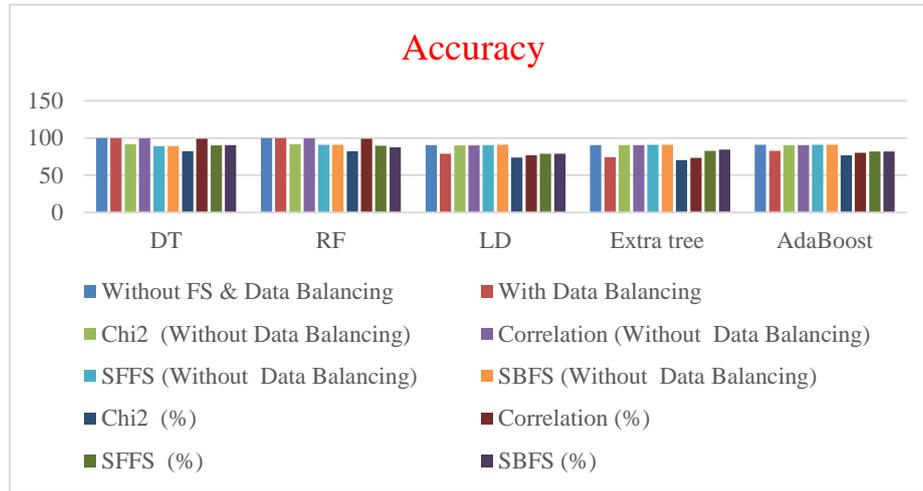

**Figure 2**: Accuracy Comparison

Table 2 represent the performance analysis for F1-Score metrics.

**Table 2.** Performance analysis of F1 Score

| ML Classifiers | Without FS &Data Balancing | With Data Balancing | Chi$^2$ (Without Data Balancing) | Correlation(Without Data Balancing) | SFFS (Without Data Balancing) | SBFS (Without Data Balancing) | Chi$^2$ | Correlation | SFFS | SBFS |
|---|---|---|---|---|---|---|---|---|---|---|
| DT | 92.12 | 91.08 | 95.18 | 92.43 | 56.18 | 46.15 | 81.19 | 90.85 | 86.58 | 86.85 |
| RF | 95.01 | 94.65 | 95.13 | 95.17 | 54.93 | 54.04 | 81.18 | 93.57 | 89.18 | 88.97 |
| LDA | 94.88 | 78.78 | 95.01 | 94.95 | 57.59 | 47.91 | 74.16 | 77.06 | 78.64 | 78.56 |
| Extra tree | 95.10 | 73.82 | 95.35 | 95.35 | 43.57 | 50.68 | 72.00 | 73.29 | 76.96 | 82.12 |
| AdaBoost | 95.18 | 82.58 | 95.32 | **95.42** | 51.63 | 52.33 | 76.23 | 79.74 | 81.47 | 81.54 |

Figure 3 illustrates the F1-score of different classifiers with data balancing, without data balancing, along with feature selection methods, or without feature selections.

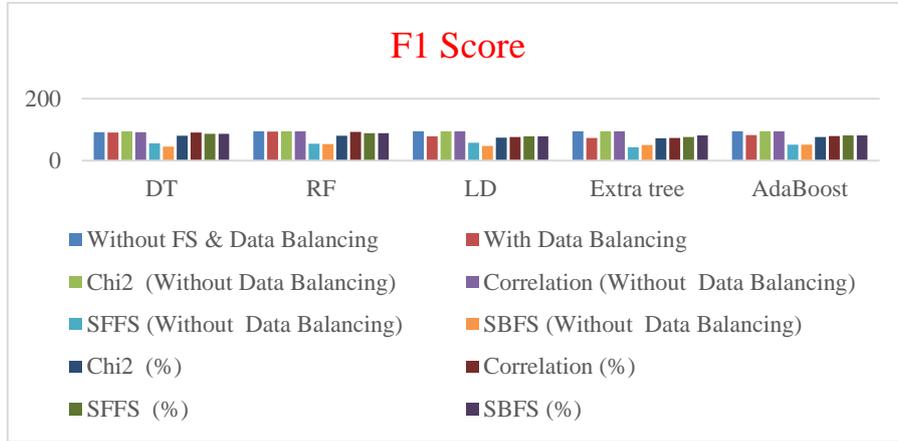

**Figure 3**: F1 score Comparison

Table 3 represent the performance analysis for Precision metrics.

**Table 3.** Performance analysis of Precision

| ML Classifiers | Without FS &Data Balancing | With Data Balancing | Chi$^2$ (Without Data Balancing) | Correlation(Without Data Balancing) | SFFS(Without Data Balancing) | SBFS (Without Data Balancing) | Chi$^2$ | Correlation | SFFS | SBFS |
|---|---|---|---|---|---|---|---|---|---|---|
| DT | 91.93 | 89.96 | 98.96 | 92.13 | 42.72 | 52.53 | 81.61 | 89.64 | 88.49 | 88.54 |
| RF | 98.78 | 98.12 | 98.88 | 98.79 | 75.23 | 77.98 | 81.41 | 95.85 | 96.41 | 94.34 |
| LDA | 97.70 | 78.75 | 97.76 | 97.48 | 45.15 | 51.58 | 75.28 | 77.13 | 77.66 | 78.21 |
| Extra tree | **99.10** | 73.85 | 98.60 | 98.10 | 64.66 | 65.38 | 77.35 | 73.80 | 80.67 | 86.79 |
| AdaBoost | 98.83 | 81.03 | 99.23 | 98.94 | 53.85 | 53.39 | 74.38 | 78.49 | 86.38 | 80.26 |

Figure 4 illustrates the precision of different classifiers with data balancing, without data balancing, along with feature selection methods, or without feature selections.

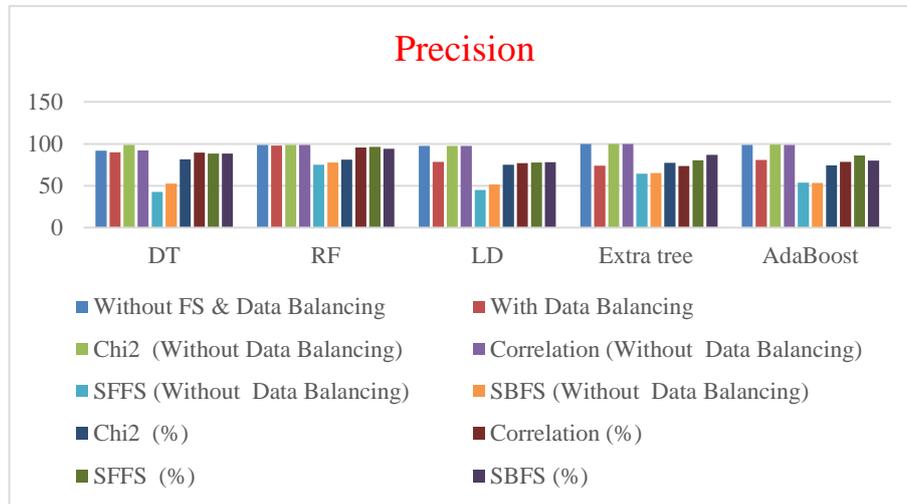

**Figure 4**: Precision Comparison

Table 4 represent the performance analysis for Recall metrics.

**Table 4.** Performance analysis of Recall

| ML Classifiers | Without FS &Data Balancing | With Data Balancing | Chi$^2$ (Without Data Balancing) | Correlation(Without Data Balancing) | SFFS(Without Data Balancing) | SBFS (WithoutData Balancing) | Chi$^2$ | Correlation | SFFS | SBFS |
|---|---|---|---|---|---|---|---|---|---|---|
| DT | 92.31 | 92.22 | 91.67 | **92.74** | 58.45 | 57.99 | 80.78 | 92.10 | 90.38 | 90.38 |
| RF | 91.51 | 91.43 | 91.65 | 91.81 | 43.60 | 47.10 | 80.97 | 91.38 | 89.64 | 89.64 |
| LDA | 92.21 | 78.80 | 92.42 | 92.55 | 59.23 | 50.19 | 73.06 | 76.99 | 79.93 | 80.50 |
| Extra tree | 90.66 | 73.78 | 91.12 | 91.12 | 59.24 | 54.41 | 67.34 | 72.78 | 78.29 | 84.06 |
| AdaBoost | 91.79 | 84.20 | 91.70 | 92.14 | 53.69 | 54.49 | 78.17 | 81.03 | 84.69 | 83.92 |

Figure 5 visually illustrates the recall of different classifiers with data balancing, without data balancing, along with feature selection methods, or without feature selections.

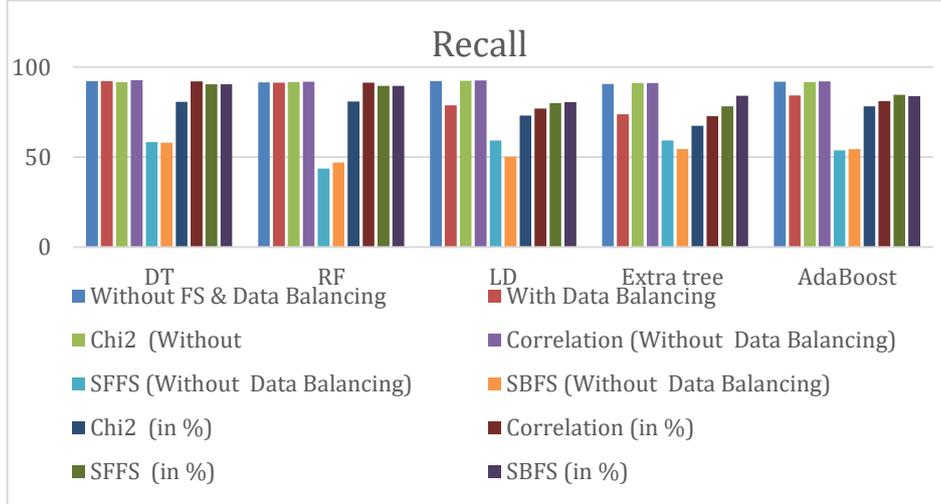

**Figure 5**: Recall Comparison

## 5      Comparative Analysis of Previous Work

We performed a comparative analysis of previous work and recorded it in tabulated form. We compared our model to previous ones, and our model produced the best results.

**Table 5.** Comparison with the previous work

| Author & Year | Method | Accuracy |
|---|---|---|
| Ashok Kumar Shanmugaraj et al. (2024) | Random Forest | 95.67 |
| Yoshihiro Kokubo et al. (2024) | Random Forest | 72 |
| K. B. M. Brahma Rao (2024) | SVC | 75 |
| Hanan Saleh Al-Messabi et al. (2024) | Logistic Regression | 83.1 |
| Daniyal Asif et al.(2023) | Extra Tree Classifier | 97.23 |
| Dhirendra Prasad Yadav (2022) | Naïve Bayes | 96 |
| Muhammad Salman Pathan et al. (2022) | Perceptron | 72 |
| L. Chandrika, K. Madhavi (2021) | HRFLM | 88.4 |
| Harshit Jindal et al. (2021) | Logistic Regression, KNN | 88.5 |
| Shu Jiang (2020) | Random Forest | 88.5 |
| Safial Islam Ayon et al. (2020) | DNN | 98.15 |
| N. Satish Chandra Reddy et al. (2019) | Random Forest | 94.96 |
| SENTHILKUMAR MOHAN et al. (2019) | HRFLM | 88.4 |
| **Proposed Method** | **Decision Tree, Random Forest (using 10-fold CV with data balancing)** | **99.83** |

## 6    Conclusion

A strategy for heart disease prediction was proposed within the confines of this investigation. To diagnose and evaluate heart disease, our study included several ML techniques, including DT, RF, Extra Tree, AdaBoost, and Linear Discriminant. The heart disease dataset utilized in this analysis raised concerns about class imbalance because the majority class had substantially more components than the minority class. K-Means SMOTE is an oversampling strategy that has been applied to overcome the oversampling problem. To increase performance, the features were selected using the Chi2 technique, Pearson correlation coefficient, SFFS technique, and SBFS technique. A method known as the 10-fold CV was employed to assess accuracy. We discovered that the decision tree and random forest classifiers provided the maximum accuracy value of 99.83 percent using data balancing when applying K-fold (10-fold) over the entire dataset.

Future research can focus on using various divisions such as 90-10, 80-20, and 70-30 of the data for training and testing parts respectively, and exploring more feature selection techniques. Furthermore, the research can be expanded to encompass additional domains of medical diagnosis and therapy, which could result in enhanced patient results.

**Acknowledgments.** We are grateful to the Computer Science Department, The University of Burdwan, India, for the continuous support from them we received while pursuing our research work.
**Disclosure of Interests.** Authors disclose that they have no conflict of interest.